\pdfoutput=1
\documentclass[11pt]{article}

\usepackage{ACL2023}

\usepackage{times}
\usepackage{latexsym}

\usepackage[T1]{fontenc}
\usepackage[utf8]{inputenc}

\usepackage{microtype}

\usepackage{inconsolata}

\usepackage{amssymb}
\usepackage{hyperref}       % hyperlinks
\usepackage{url}            % simple URL typesetting
\usepackage{booktabs}       % professional-quality tables
\usepackage{amsfonts}       % blackboard math symbols
\usepackage{nicefrac}       % compact symbols for 1/2, etc.
\usepackage{microtype}      % microtypography
\usepackage{xcolor}         % colors
\usepackage{graphicx}
\usepackage{amsmath} 

\title{Multi-View Zero-Shot Open Intent Induction from Dialogues:\\ Multi Domain Batch and Proxy Gradient Transfer}

\author{
  Hyukhun Koh$^{1}$, Haesung Pyun$^{2}$, Nakyeong Yang$^{1}$, Kyomin Jung$^{1}$   \\
  Seoul National University$^{1}$, Hanyang University$^{2}$   \\
 \texttt{\{hyukhunkoh-ai, yny0506, kjung\}@snu.ac.kr}   \\
  \texttt{ haesung.pyun@gmail.com}
}

\begin{document}
\maketitle
\begin{abstract}
In Task Oriented Dialogue (TOD) system, detecting and inducing new intents are two main challenges to apply the system in the real world. In this paper, we suggest the semantic multi-view model to resolve these two challenges: (1) SBERT for General Embedding (GE), (2) Multi Domain Batch (MDB) for dialogue domain knowledge, and (3) Proxy Gradient Transfer (PGT) for cluster-specialized semantic.
MDB feeds diverse dialogue datasets to the model at once to tackle the multi-domain problem by learning the multiple domain knowledge.
We introduce a novel method PGT, which employs the Siamese network to fine-tune the model with a clustering method directly.
Our model can learn how to cluster dialogue utterances by using PGT. 
Experimental results demonstrate that our multi-view model with MDB and PGT significantly improves the Open Intent Induction performance compared to baseline systems.
\end{abstract}

\section{Introduction}

\; In developing Task Oriented Dialogue (TOD) systems, it is important to understand the user's intent, such as opening a new account in banking and reporting accidents in finance. In this circumstance, TOD system is used to capture the intents automatically within some dialogue turns, generating appropriate responses to users. Since Pretrained Language Models (PLMs) have become mainstream among the recent NLP tasks, various kinds of research to apply the PLMs to a dialogue system have appeared. (\citet{henderson-etal-2019-repository}, \citet{bao-etal-2020-plato}, \citet{majumder-etal-2020-interview}, \citet{zhao-etal-2020-knowledge-grounded}) 

In TOD systems, one of the core tasks is intent detection, which seeks to discern user intentions from their utterances. However, most of the typical intent detection modules depend heavily on the domain to be applied. Due to the domain-lopsided characteristic, it is necessary for typical models to collect the data of the domain to be serviced and go through a cumbersome process to train a model for each domain from scratch. Besides, no standard conventions exist for defining and labeling the intents within the same domain.
For example, Banking77, an intent classification dataset, has 77 intents. In contrast, Banking from DSTC 11 has 29 intents and some of the intents, "AskAboutATMFees" and "GetBranchInfo," are not included in Banking77. Hence, it is tricky to apply a model trained with Banking77 directly to Banking dataset.
For example, Banking77, an intent classification dataset, has 77 intents. In contrast, Banking from DSTC 11 has 29 intents and some of the intents, "AskAboutATMFees" and "GetBranchInfo," are not included in Banking77. Hence, it is impossible to directly apply a model trained with Banking77 to Banking dataset. 
In addition, customers may not communicate within only predefined intent sets. Summing up, intent detection modules should induce intents from utterances beyond each domain dataset.

Previous researches utilized BERT to produce hidden representations and passed them to clustering algorithms in order to discern intents of utterances (\citealp{DBLP:journals/corr/abs-1911-08891}, \citealp{aharoni-goldberg-2020-unsupervised}). However, utterances in dialogue data are usually shorter and colloquial compared to generic text where the ellipsis and omission are less likely to occur. Furthermore, BERT representations are insufficient to capture the intent information for dialogue since it is only trained to focus on learning the context of the generic text. Additional methods are required to contain the domain-specific information since the meaning of a sentence can also be interpreted in various ways depending on its domain.
Therefore, some researchers have trained PLMs with dialogue data (\citealp{shen-etal-2021-semi}, \citealp{Zhang_Xu_Lin_Lyu_2021}, \citealp{zhang-etal-2021-effectiveness-pre}, \citealp{zhang-etal-2021-shot}). However, they mostly have conducted experiments within a few-shot setting where some portion of a target dataset is available, rather than a zero-shot setting where target domain data is not accessible.
In accordance with the study by \citet{zhang-etal-2022-new}, they handled such problems in semi-supervised learning schemes. \citet{zhang-etal-2022-new} showed that it is beneficial to train PLM with external intent detection datasets. They also proposed a contrastive learning method based on K-nearest neighbor algorithm. Nevertheless, the performance gap between few-shot and zero-shot settings has been still large.

Our whole pipeline is the same as the baseline used in DSTC 11 task2\footnote{\href{https://drive.google.com/file/d/1itlby2Ypq3sRVtOY1alr3ygjPZZdB2TT/view}{DSTC11 Track Proposal: Intent Induction from Conversations for Task-Oriented Dialogue}}. Details are presented in Section \hyperref[task description]{4 Task Description}. Following the baseline pipeline, we only feed some utterances with labeled as InformIntent without any modification. Similar utterances should be clustered in the hidden representation space.

In this paper, we aim to improve Open Intent Induction which requires inducing intent sets in various domains without additional training on target domain datasets (zero-shot setting). We suggest the semantic multi-view model to resolve aforementioned problem using three methods : (1) SBERT for general embedding (GE), (2) Multi Domain Batch (MDB) for dialogue domain knowledge, and (3) Proxy Gradient Transfer (PGT) for cluster-specialized semantics. By MDB and PGT methods, constructed domain-agnostic model is not necessary to update parameters further when applying to a different target domain. MDB makes a model to learn various domain knowledge by composing a batch with six groups of samples, sampling from each of the six different dataset. PGT is a way of fine-tuning with K-means to improve on the clustering capability, especially in inducing intents properly. Although K-means is an integer assignment problem that is impossible to differentiate, PGT allows a differentiable comparison of K-means labels and ground truth labels by adopting Siamese network learning process.

From our experiment, combining the General Embedding (GE) module and the MDB module showed an excellent performance. In addition, when applying the PGT method to these two modules for the clustering, the performance was improved as well. Finally, we found that Spectral clustering with all three modules (GE, MDB, PGT) works better, compared to K-means clustering. The codes are available in 
the github repository\footnote{\href{https://github.com/hyukhunkoh-ai/multi_view_zero_shot_open_intent_induction.git}{
Multi\_View\_Zero\_Shot\_Open\_Intent\_Induction}}

This paper has three contributions:
\begin{enumerate} 
   \item We suggest the multi-view model composed with GE, MDB, PGT modules.
  \item We suggest a novel method PGT to directly fine-tune a model onto a clustering. PGT contributes to better performance and deals with a non-differentiable problem.
  \item Our model demonstrates excellent performance in Banking and Finance, outperforming the baseline in Open Intent Induction task.
\end{enumerate}

\section{Background}

\;  In this section, we will briefly cover PLMs (\ref{PLMs}) and explore how PLMs are applied to TOD systems (\ref{OODs}). Finally, we will introduce existing clustering methods (\ref{clustering}).

\subsection{Pretrained Language Models}
\label{PLMs}
\; There are two streams of PLM which is trained on massive corpus, Auto-Regressive (AR) and Auto-Encoding (AE) model. 
GPT-3 \citep{DBLP:journals/corr/abs-2005-14165} and FLAN \citep{wei2022finetuned} are well known for AR model using Transformer decoder which are optimized to maximize the likelihood of the next word generation given preceding tokens. 
In comparison, BERT  \citep{devlin-etal-2019-bert}, a representative of AE model, is a pretrained Transformer encoder with Masked Language Modeling (MLM).
BERT variant models have been introduced since BERT has shown an outstanding performance in Natural Language Understanding (NLU) tasks. RoBERTa \citep{liu2020roberta} enhanced BERT with Dynamic Masking method and larger dataset. Sentence Transformer SBERT \citep{reimers-gurevych-2019-sentence} derives more semantically meaningful sentence embeddings by training a model with NLI task in the scheme of Siamese network.

Some studies tried to overcome the disadvantage of AR and AE model by combining MLM with permutation methods: XLNet \citep{NEURIPS2019_dc6a7e65}, MPNet \citep{NEURIPS2020_c3a690be}. XLNet \citep{NEURIPS2019_dc6a7e65} proposed an autoregressive pretraining by permutating tokens to capture long dependencies. MPNet unifies MLM and permutation pretraining methods, employing a two-stream attention to capture position information.

\subsection{Dialogue Systems with PLMs}
\label{OODs}
\; Despite its universal applications of PLMs, there is a discrepancy between a general corpus and a dialogue corpus. Therefore, there are some studies which trained the PLMs on a dialogue corpus further, rather than directly applying PLMs to the TOD task. TOD-BERT \citep{wu-etal-2020-tod} pretrained BERT on TOD corpus. IntentBERT \citep{zhang-etal-2021-effectiveness-pre} proposed regularized supervised learning and showed that it is promising in a few-shot intent detection task. \citet{Vulic2021ConvFiTCF} proposed two-stage procedures. They used a universal conversational encoder with small conversational data samples, subsequently fine-tuned an encoder for a sentence similarity task. These studies focus on pretraining with the dialogue corpus and applying it to a specific dataset.
\subsection{Clustering Methods}
\label{clustering}
\; There are several candidates when cluster task is necessary.
K-means \citep{MacQueen1967SomeMF} clusters the inputs based on Euclidean distance. It has problems when the real clusters do not have spherical shape or have a huge imbalance in size. Furthermore, Clustering performance heavily relies on the predefined K.
Agglomerative \citep{CHIDANANDAGOWDA1978105}, one of hierarchical clusterings which build clusters based on the proximity matrix, starts with N clusters. 
However, it is often deficient in robustness, sensitive to noise, and computationally expensive. 
DBSCAN \citep{10.5555/3001460.3001507} is a density-based approach, which finds every point’s neighbors and identifies the core points. DBSCAN is robust to noise and outliers. Nevertheless, it is slower than K-means clustering. It is also difficult to choose distance threshold epsilon and minimum points. ITER-DBSCAN \citep{chatterjee-sengupta-2020-intent} extends DBSCAN with label propagation technique and dialog act classification.
Gaussian Mixture Model (GMM) clusters based on the probability of a mixture of multiple unknown Gaussian distributions. This method does not assume the real cluster to be a specific shape but requires high computational costs.
Spectral clustering \citep{NIPS2001_801272ee} is based on Ratio Cut algorithm using the Laplacian matrix. This algorithm calculates the Laplacian matrix upon the similarity matrix and obtains eigenvectors from the matrix. This method can cluster the data type of non-global shape. Nonetheless, the number of clusters is essential in advance.

\section{Related Works}
\;  Studies related to OOD detection and intent discovery are very relevant to Open Intent Induction task in terms of the fact that they need to handle unseen intents. The researchers of those studies are interested in how to make a good representation and how to connect it well by using a grouping algorithm. \citet{perkins-yang-2019-dialog} proposed the method which utilized content view and utterance view to yield the cluster assignments.  \citet{lin-xu-2019-deep} introduced margin loss on Bi-Directional Long Short Term Memory (BiLSTM) \citep{6998838} for inter-class discrimination and intra-class compactness. They applied local outlier factor (LOF) \citep{10.1145/342009.335388} to detect unknown intents based on the nearest-neighbor's density. \citet{chatterjee-sengupta-2020-intent} introduced ITER-DBSCAN which extended DBSCAN with a label propagation technique. However, the methods mentioned above require domain specific knowledge and laborious feature engineering. Also, those methods often cannot fully leverage the prior knowledge and cannot accurately capture fine-grained intents.
\citet{Lin_Xu_Zhang_2020} leveraged PLM and limited labeled data for supervised intent classification. They used the sentence similarity and filtered out low confidence assignment which is for refining clusters. Deep Aligned Clustering \citep{Zhang_Xu_Lin_Lyu_2021} pretrained BERT model with known intent samples to inject prior knowledge. As a next step, they produced pseudo-labels by the cluster assignment for self-supervised learning. 
CPFT \citep{zhang-etal-2021-shot} performed a contrastive learning with MLM to learn discriminative ability on utterance semantics. \citet{shen-etal-2021-semi} suggested the supervised contrastive scheme with dialogue domain labeled dataset on the basis of MPNet. After that, they applied K-means clustering with estimated K based on Bayesian Optimization.  
\citet{zhou-etal-2022-knn} focused on training fine-grained discriminative semantic features to deal with OOD intents among in-domain (IND) distribution. They utilize K-nearest neighbors of IND intents sample.
The limitation of those studies is that they mainly focused on a few-shot performance where the premise is that few target-related labels are available. 

\citet{chen-etal-2022-intent} pretrained Transformer with dialogue data and introduced a domain adaptive pretraining method which shows good performance in Open Intent Discovery task. Since their domain adaptive pretraining method shows promising results, we adopt it with some variations. 
\citet{zhang-etal-2022-new} embedded rich information for the clustering by exploiting external and internal data. They show comparable results in the setting where target-related labels are unprovided. However, there is a still big gap between a few-shot setting and a zero-shot setting.

In this paper, we suggest two learning methods to induce intents from dialogues efficiently: 1. For MDB, how to appropriately inject prior external knowledge in PLM model to be robust whenever the domain changes and 2. For PGT, how the capability of model can be boosted in terms of clustering performance. Using two methods, the performance of Banking and Finance significantly increased compared to the model exposed in a few-shot setting.

\section{Task Description}\label{task description}

\begin{figure}[h] %%% t: top, b: bottom, h: here
\centering
    \includegraphics[width=1\linewidth]{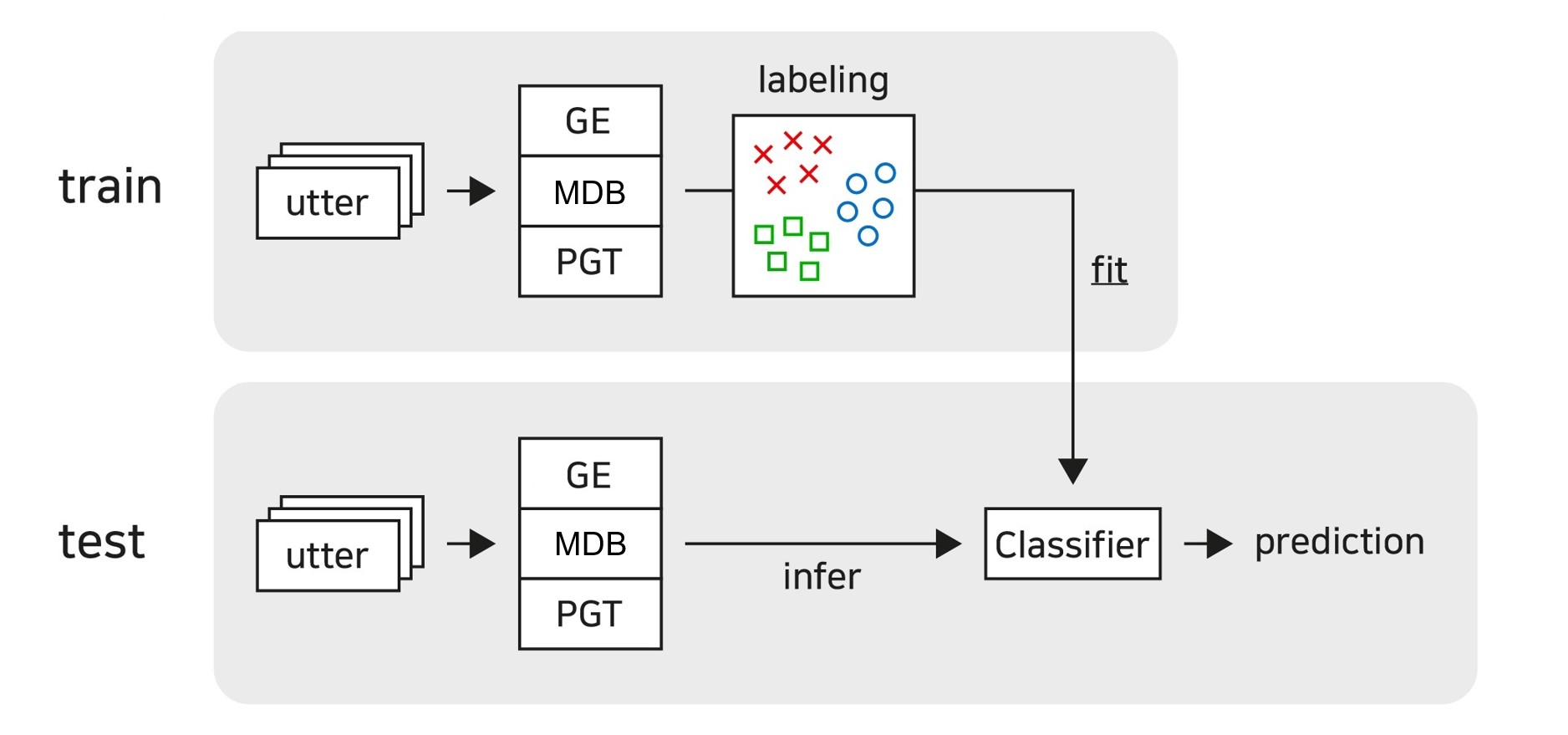}
    \hfil
\caption{This is the outline of DSTC 11 task 2 Open Intent Induction.
By concatenating representations of GE, MDB, and PGT modules, a clustering algorithm clusters those multi-view representations of train-utterances to construct an intent schema which consists of utterances and predicted labels. Next, we train the linear classifier with the intent schema. At test phase, multi-view representations of test-utterances go into the trained classifier to induce labels.
}
\label{pipeline}
\end{figure}
\;  The second track of DSTC11 challenge is Intent Induction task based on TOD, such as Banking and Finance. This challenge aims to automatically induce intent sets for a task-oriented chatbot without training on target datasets which is a customer service interaction between the chatbot and clients.

There are three given datasets(Insurance, Banking, Finance) from 1K customer support spoken conversations, Insurance for development and others for evaluation. Each dataset has two divided data, one for training an evaluation classifier and the other for testing the performance of Open Intent Induction. The challenge assumes that utterances have only speaker roles without labels such as dialog act and intents, the number of which ranges from 5 to 50.

The task consists of three subtasks: Label Construction, Classifier Construction, and Inducing Intent. Label Construction subtask clusters the train-utterances in order to make a set of intents while predicting what the number of intents is. Classifier Construction subtask is for training a classifier with train-utterances and predicted labels which are from Label Construction subtask. Inducing Intent subtask predicts the labels for test-utterances based on a trained classifier.

\subsection{Problem Formulation}

\begin{equation} \label{eq1}
\begin{split}
  \left\{ \right.d_1 &\cdots d_n \left. \right \} \rightarrow  \left\{ \right.u_1 \cdots u_m \left. \right \}, \\
   & d_i \subset D \; and \; u_i \subset U 
\end{split}
\end{equation}
% in other words, some utterances labeled as InformInent, 추가
According to equation (\ref{eq1}), we denote a dialogue set as $\mathbf{D = \{d_1 \cdots d_n\}}$, where $\mathbf{d_i}$ represents the i-th utterance in a given dialogue, n is the total number of the dialogue sets. Extracted utterances denote as $\mathbf{U = \{u_1 \cdots u_m\}}$, where $\mathbf{u_i}$ represents the i-th candidate utterance which seems to have intent, labeled as InformIntent, and m is the number of utterances selected from $\mathbf{D}$. Each single turn utterance becomes an input to the model and $\mathbf{U}$ may have some noises containing utterances without intents. 

For Label Construction subtask, K-means attaches labels to train-utterances on hidden representation spaces that our model(GE-MDB-PGT) generates. For Classifier Construction subtask, train-utterances and their labels are used to train Linear Regression classifier. For Inducing Intent subtask, test-utterances sequentially go into our model and trained classifier to induce intents. The whole pipeline of DSTC 11 task 2 is in Figure \ref{pipeline}.

\section{Methods}\label{Methods}
\begin{figure}[h] %%% t: top, b: bottom, h: here
\centering
    \includegraphics[width=1\linewidth]{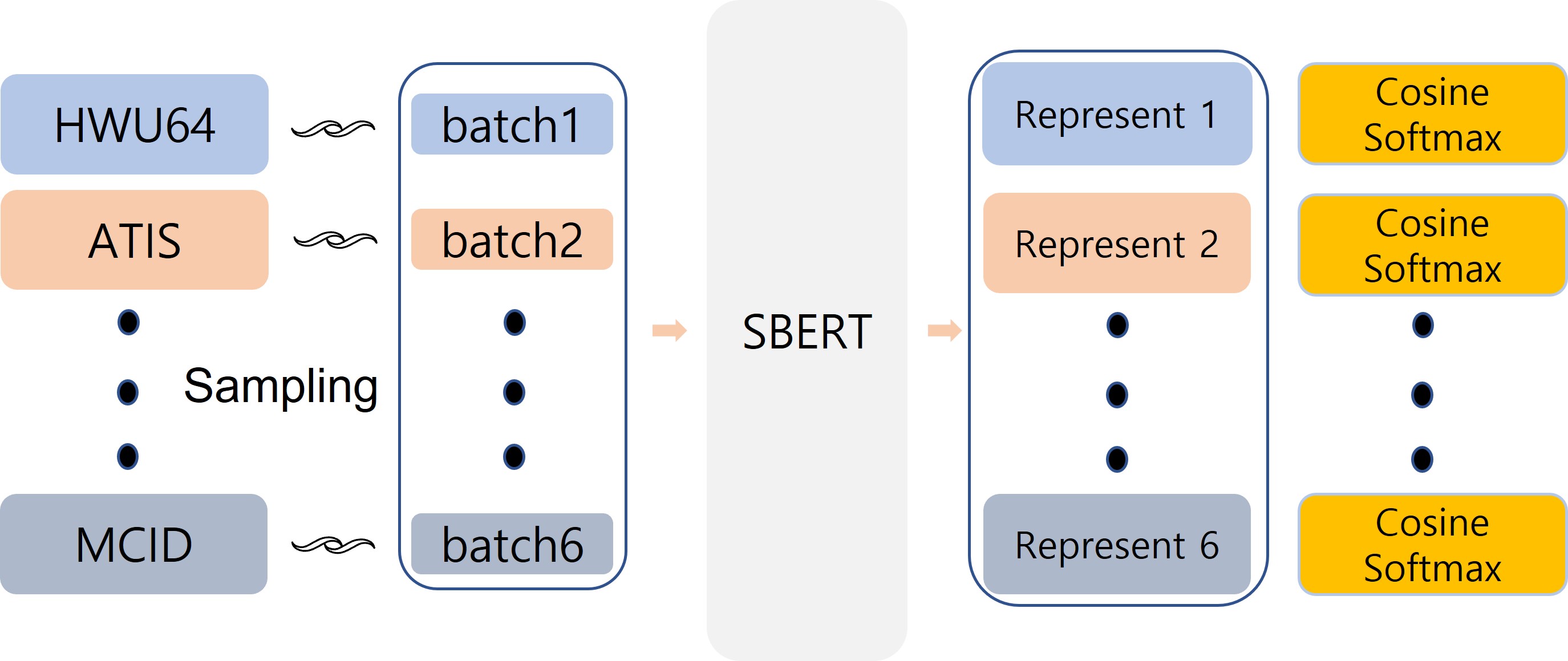}
    \hfil
\caption{It shows the procedure of MDB to learn how to make appropriate domain-agnostic hidden representations. Multi-domain samples from each dataset construct the total batch (HWU64, ATIS, ..., MCID). Next, SBERT converts the total batch to hidden representations. Finally, the loss is calculated by applying cosine softmax for samples of each dataset.}
\label{MDB}
\end{figure}

\; In this section, we cover our methodology in detail. We adopt the baseline model structure of the DSTC 11 challenge as it is, but we investigate how the performance can be boosted by further learning. Our goal is for the model to automatically induce and generate meaningful intents from a collection of unlabeled utterances. SBERT, the baseline model proposed by DSTC 11, is used as the backbone.

\subsection{Multi-View Model}
 \;  Utterances from $\mathbf{U}$ enter into our three modules for taking the different perspectives into account, which is a multi-view approach. By passing U through them, we get each of three hidden representations of utterance.
\begin{equation} \label{eq2}
\begin{split}
\centering
   Z_1 &= GE(U)  \\
   Z_2 &= MDB(U) \\
   Z_3 &= PGT(U)    \\
   H = C&oncat\left\{ Z_1,Z_2,Z_3 \right\}
\end{split}
\end{equation}
Hidden representation $Z_1$ is drawn using the General Embedding(GE) module, which interprets utterances in terms of a general linguistic perspective. $ Z_2$ is created through a module learned by MDB to extract the dialogue-viewed representation. Hidden representation $Z_3$ is induced by the PGT module for tuning an encoder and a clustering module in order to boost the performance. H is constructed by concatenating all those hidden representations.

In the last step, K-means or Spectral clustering groups the concatenated representations to make intent schema by Bayesian Optimization (BO). BO method samples the number of clusters from a uniform distribution between 5 to 50 and estimates the optimal K based on the silhouette score. In the test process, Hungarian algorithm maps predicted labels to reference labels in order to evaluate the performance.

\subsection{Multi Domain Batch Cosine Softmax (MDB)}

\;  As mentioned earlier, SBERT (the baseline of DSTC 11) is not enough to make a proper representation of the dialogue domain due to the discrepancy between the generic corpus and the dialogue corpus. Therefore, training PLM with a dialogue corpus is indispensable. In that sense, Multi Domain Batch (MDB) proved to be efficient in \citet{chen-etal-2022-intent}. Unlike \citet{chen-etal-2022-intent}, we directly apply MDB to PLM with some variations. The Figure \ref{MDB} explains the process step by step.
MDB uses six external datasets which are publicly available. 
In the process of MDB training, six different dialogue datasets configure the batch. For example, if the batch size is 36, the model would convert six samples for one dataloader to hidden representations where samples for one dataset are independent of the other datasets. The last layer of MDB is cosine softmax which is already proven to be more advantageous than naive softmax in \citet{chen-etal-2022-intent}. We add a term $\tau$ for a numerical and denominator part.
\begin{equation} 
\centering
\small
j^{th} \; Cosine \; Softmax = \frac{exp(\bar{h_j}^T\bar{w^k}/\tau )}{\sum_{i}^{L^{k}}exp(\bar{h_i}^T\bar{w^k}/\tau )} ,
\end{equation} 
where k is the $k^{th}$ dataset, $\bar{h_i}$ for 
$h_i$ divided by norm($h_i$), $\bar{w^{k}}$ for $k^{th}$ dataset’s linear normalized weight, $\tau $ for the temperature, and $L^{k}$ for the $k^{th}$ dataset's number of intent categories.

The benefit of this methodology is that we do not have to preprocess collected datasets to unify the intent category sets. It saves lots of time to merge the datasets because different domains have different intent label categories, and even the same domain has different ways of defining intent labels. Next, the model can acquire multiple dialogue domain knowledge at once, which is not dependent on a specific dataset. It indicates that we no longer train the model further when applying the model to other TOD industrial fields. Besides, as well known fact that catastrophic forgetting could occur when sequentially training the model, MDB prevents this problem by parallelly training which is much more efficient in terms of learning time.

\begin{figure*}[t] %%% t: top, b: bottom, h: here
\begin{center}

\includegraphics[width=1\linewidth]{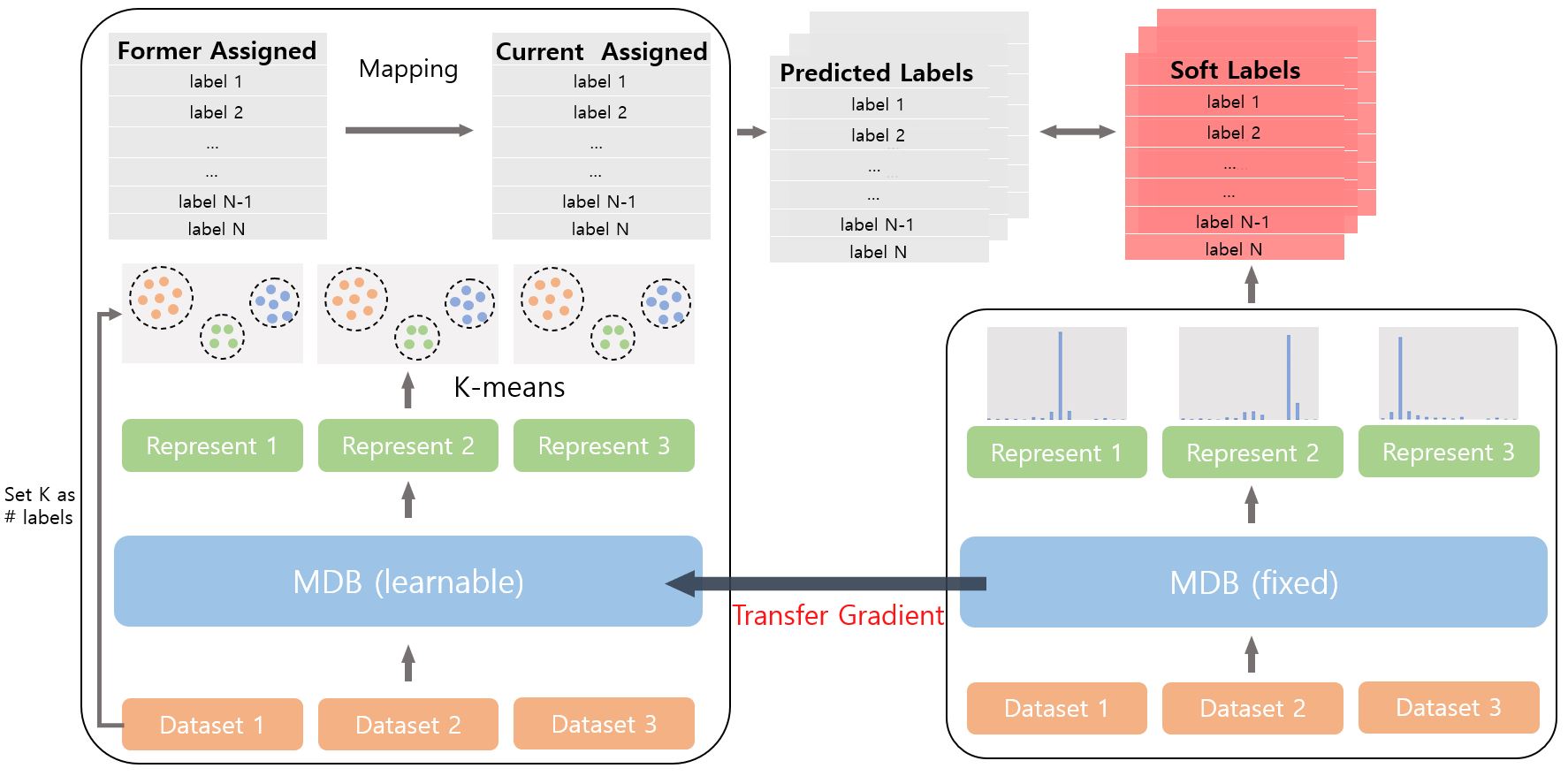}

\end{center}
\caption{MDB signifies the SBERT trained by MDB method. First, two same MDB are initialized. Second, MDB (learnable) predicts the labels for train utterances using K-means. Third, MDB (fixed) calculates gradients based on cross entropy loss between predicted labels and soft labels. Fourth, MDB (fixed) transfers the gradients to MDB (learnable), considering a weight difference. Finally, MDB (learnable) is optimized with transferred gradients.}
\label{PGT}
\end{figure*}
\subsection{Proxy Gradient Transfer(PGT)}

\; In Open Intent Induction task, the model induces a set of intents by clustering methods. Therefore, the model needs to know how to cluster well from the perspective of clustering methods. In that sense, Proxy Gradient Transfer(PGT) is a way of learning how to interpret utterances in terms of clustering methods, grounded on the MDB module. Since K-means self-supervised learning is proven to be efficient in \citet{Zhang_Xu_Lin_Lyu_2021}, we choose K-means as a clustering module. Details are in Figure \ref{PGT}. However, there are three obstacles to overcome. It is not straightforward to select the K for K-means and it is impossible to differentiate. Furthermore, K-means is non-convex, whose result highly depends on fortunate initialized centroids. At last, the labels of clusters can vary each step even though their contents do not change.

First, we use the number of intent labels for each dataset to determine K. The premise is that K does not change during PGT training. 
Second, Siamese network learning process makes the model differentiate. In the process of PGT training, one of two identical encoders in Siamese network (MDB-learnable) predicts labels from K-means clustering. The other (MDB-fixed) generates representations to use them as soft labels and vicariously calculates gradients based on the cross entropy loss between predicted labels and soft labels. Afterward, the gradients are assigned from MDB(fixed) to MDB(learnable), reflecting equation (\ref{transfer}). 
\begin{equation}
\centering
\label{transfer}
\small
Grad_{learned} = Grad_{fixed} + W_{learned} - W_{fixed}
\end{equation}
where $Grad_{learned}$ is the gradient of MDB (learnable), $Grad_{fixed}$ for the gradient of MDB (fixed), $W_{learned}$ for the weight of MDB (learnable), and $W_{fixed}$ for the weight of MDB (fixed).
In the next step, MDB(learnable) is optimized with allotted gradients. 
Third, when the performance of MDB is reasonably good, the effect of initialized centroids diminishes. Though few initial randomnesses can still remain, we consider it as an augmentation. Lastly, the cluster label is adjusted through Hungarian mapping according to the similarity among the clusters' centroids.

\section{Experiment}

\subsection{Dataset}
% We split  
\; We separate our datasets into two groups, a train dataset and a test dataset. For MDB and PGT training, we introduce six datasets.

\textbf{Train} Banking77 \citep{casanueva-etal-2020-efficient} is banking dialogue divided into Out-Of-Domain (OOD) intent and In-Domain (IND) intent. We use fine-grained 77 intents as train data. ATIS \citep{ hemphill-etal-1990-atis} is made of airplane travel transcripts. We preprocess them to extract the 17 intent categories. Clinc150 \citep{ larson-etal-2019-evaluation} is for evaluating the intent detection system. HWU64 \citep{DBLP:journals/corr/abs-1903-05566} is for developing a home robot with various scenarios to serve as personal assistance, comprised of multi-domain(21 domains). MCID \citep{levy-wang-2020-cross} is involved in Covid-19, a chatbot dataset consisting of user and agent. Restaurants-8k \citep{coope-etal-2020-span} is a set of 8,198 utterances gathered from the actual restaurant booking system. The details of the training dataset with specific data statistics are in table~\ref{train-dataset}. 

\textbf{Test} We test our model on two different domain datasets without further training. Finance and Banking are the released test datasets of DSTC 11.

\begin{table*}
    \centering
    \caption{Intent Detection Datasets Summary}
    \label{train-dataset}

    \begin{tabular}{|l|l|l|l|l|l|}
    \toprule
    Dataset  & \#Domain & \#Intents & \#Utters & Vocab & Length(min / max / avg)  \\
    \midrule
    ATIS           & 1        & 25        & 5,781        & 938        & 1 / 46 / 11.2            \\
    BANKING77      & 1        & 77        & 13,072       & 2,636      & 2 / 78 / 11.7            \\
    CLINC150       & 10       & 150       & 22,500       & 6,420      & 1 / 28 / 8.3             \\
    HWU64          & 21       & 64        & 11033        & 4661       & 1 / 25 / 6.6             \\
    MCID           & 1        & 16        & 1745         & 1415       & 1 / 20 / 6.7            \\
    RESTAURANT8K   & 1        & 13        & 4181         & 3949       & 1 / 84 / 15.1            \\
    \bottomrule
\end{tabular}
\end{table*}

\subsection{Baseline}

\; In DSTC 11 task2 Open Intent Induction, "glove-840b-300d" and "sentence-transformer/all-mpnet-base-v2" are base encoders. Afterward, K-means clusters the hidden representations from encoders to induce an intent schema. Utterances with "InformIntent" are used as input, which implies that some noises might be included as well.

\begin{itemize}
 \item  KGlove + K-means clustering
  \item SBERT + K-means clustering
\end{itemize}

\subsection{Metrics}

There are five evaluation metrics: accuracy (ACC), normalized mutual information (NMI), Adjusted rand index (ARI), Precision, Recall, and F1 score. ACC is the main evaluation. NMI is defined as 
\begin{equation}
    \mathbf{NMI(Y,C)} = {\frac{2*I(Y;C)}{[H(Y) + H(C)]}},
\end{equation}
where Y is class labels, C for cluster labels, H for entropy, and I for mutual information. Following the baseline code, we used hyperopt packages to identify the optimal alignment.

\subsection{Implementation}
\; The "sentence-transforemr/all-mpnet-base-v2" model is our backbone. In MDB training process, we split each dataset into three sets (Train 70 : valid 20 : test 10). Since each dataset has its own various sizes, we set the number of iterations per epoch according to the largest dataset's length (i.e., 22,500 of CLINC150 dataset) divided by the batch size 64. 

We train MDB and PGT modules 40 epochs in total. The optimizer is AdamW with cosine annealing and warm-up scheduler, whose learning rate is 5e-6, weight decay for 1e-2, and warm-up steps for 10\% of total iterations. To calculate the loss in the MDB training, we set temperature $\tau$ as 0.05 for each dataset, whereas PGT uses cross entropy loss.

In our experiments with the grouping algorithms, K-Means and Spectral clustering use n init 10. As mentioned in Section \hyperref[Methods]{5 Methods}, we sample from the uniform distribution between 5 and 50 until finding optimal K. The type of affinity is {'nearest neighbors'} for Spectral clustering.

\section{Results}
\label{results}
\begin{table*}[]

\setlength{\tabcolsep}{3pt}
\centering
\footnotesize
\begin{tabular}{|c|ccc|ccc|ccc|}
\hline
               & \multicolumn{1}{c}{}                  & \multicolumn{1}{c}{Banking}           & \multicolumn{1}{c|}{}     & \multicolumn{1}{c}{}                  & \multicolumn{1}{c}{Finance}           & \multicolumn{1}{c|}{}      & \multicolumn{1}{c}{}                  & \multicolumn{1}{c}{Summary}           & \multicolumn{1}{c|}{}       \\
Models                      & \multicolumn{1}{c}{ACC}               & \multicolumn{1}{c}{NMI}               & \multicolumn{1}{c|}{F1}   & \multicolumn{1}{c}{ACC}               & \multicolumn{1}{c}{NMI}               & \multicolumn{1}{c|}{F1}    & \multicolumn{1}{c}{ACC}               & \multicolumn{1}{c}{NMI}               & \multicolumn{1}{c|}{F1}    \\ \hline
Baseline (TFF-kmeans)       & \multicolumn{1}{c|}{70.76}            & \multicolumn{1}{c|}{83.97}            & \multicolumn{1}{c|}{79.9} & \multicolumn{1}{c|}{56.46}            & \multicolumn{1}{c|}{76.22}            & \multicolumn{1}{c|}{67.83} & \multicolumn{1}{c|}{63.61}            & \multicolumn{1}{c|}{80.10}           & \multicolumn{1}{c|}{73.87} \\ \hline
FTF - kmeans                & \multicolumn{1}{c|}{74.45}          & \multicolumn{1}{c|}{87.33}           & 82.54                   & \multicolumn{1}{c|}{70.18}           & \multicolumn{1}{c|}{82.72}           & 78.12                    & \multicolumn{1}{c|}{72.31}          & \multicolumn{1}{c|}{85.02}          & 80.33                    \\
FTF - spectral              & \multicolumn{1}{c|}{83.78}          & \multicolumn{1}{c|}{89.43}          & 86.40                    & \multicolumn{1}{c|}{68.58}          & \multicolumn{1}{c|}{83.89}          & 78.79                    & \multicolumn{1}{c|}{76.18}         & \multicolumn{1}{c|}{86.66}          & 82.59                     \\ \hline
FTT - kmeans                & \multicolumn{1}{c|}{75.18}          & \multicolumn{1}{c|}{85.82}          & 81.52                   & \multicolumn{1}{c|}{65.22}          & \multicolumn{1}{c|}{80.64}          & 74.07                    & \multicolumn{1}{c|}{70.20}         & \multicolumn{1}{c|}{83.23}          & 77.80                    \\
FTT - spectral              & \multicolumn{1}{c|}{87.22}          & \multicolumn{1}{c|}{91.17}          & 89.73                   & \multicolumn{1}{c|}{70.177}           & \multicolumn{1}{c|}{84.25}          & 79.67                    & \multicolumn{1}{c|}{78.70}          & \multicolumn{1}{c|}{87.71}         & 84.70                    \\ \hline
TTF - kmeans                & \multicolumn{1}{c|}{79.36}           & \multicolumn{1}{c|}{87.71}           & 83.66           & \multicolumn{1}{c|}{57.52}           & \multicolumn{1}{c|}{78.44}           & 70.38                    & \multicolumn{1}{c|}{68.44}         & \multicolumn{1}{c|}{83.08}           & 77.02                    \\
TTF - spectral              & \multicolumn{1}{c|}{89.68}          & \multicolumn{1}{c|}{92.66}          & 92.30                   & \multicolumn{1}{c|}{70.62} & \multicolumn{1}{c|}{\textbf{84.97}} & \textbf{80.20}                    & \multicolumn{1}{c|}{80.15}         & \multicolumn{1}{c|}{\textbf{88.81}}          & \textbf{86.25}                    \\ \hline
TTT - kmeans                & \multicolumn{1}{c|}{74.20}            & \multicolumn{1}{c|}{86.98}           & 82.10            & \multicolumn{1}{c|}{60.8}            & \multicolumn{1}{c|}{80.48}           & 73.73                     & \multicolumn{1}{c|}{67.5}         & \multicolumn{1}{c|}{83.73}          & 77.92                    \\
TTT - spectral              & \multicolumn{1}{c|}{\textbf{90.17}} & \multicolumn{1}{c|}{\textbf{92.99}} & \textbf{92.68}          & \multicolumn{1}{c|}{\textbf{71.06}}          & \multicolumn{1}{c|}{84.20}          & 79.74           & \multicolumn{1}{c|}{\textbf{80.62}} & \multicolumn{1}{c|}{88.60} & 86.20         \\ \hline
\end{tabular}
\caption{Results represent the performance of our method applied to Banking and Finance domain datasets. We notate T(True) and F(False) to signify each module’s existence (GE-MDB-PGT). FTF refers to a model using MDB, FTT refers to a model using  MDB and PGT, TTF refers to a model using GE and MDB, and finally TTT refers to a model using GE, MDB and PGT}
\label{result_table}
\end{table*}

\; We show the results on the Banking and Finance DSTC 11 datasets in Table \ref{result_table}.
We ablate each module to find the best combination. We notate T(True) and F(False) to signify each module's existence (GE-MDB-PGT). For example, TTF-kmeans means that we concatenate representations of GE and MDB, subsequently clustering those with K-means algorithm.

First, an encoder with Spectral clustering outperforms one with K-Means clustering in most cases by at least around 5\% in ACC and at most 16\% in ACC. Spectral clustering additionally constructs an affinity matrix using the similarity based on the nearest neighbors and clusters utterances based on K-means in a Laplacian embedding space. Therefore, the result implies that Spectral clustering is advantageous for Finance and Banking as it clusters data by considering the connectivity between data points rather than compactness around the cluster center.
More details are in Appendix \ref{Banking appendix} and Appendix \ref{Finance appendix}

Next, the MDB module significantly plays a significant role in Open Intent Induction task beyond the domain. For instance, on Banking dataset, FTF-spectral outperforms Baseline by around 13\% in ACC, 5\% in NMI, and 7\% in F1 score. On Finance Dataset, it also shows higher performance, around 12\% in ACC, 7\% in NMI, and 11\% in F1 score. These results symbolize that MDB method is significantly profitable for learning how to discriminate similar utterances without a domain dependency. Additionally, by observing TTF model's performance improvement upon Baseline (TTF-kmeans), we can derive the same result.

Furthermore, comparing FTF and FTT, the PGT module boosts performance by bridging the gap between clustering and our model. On Banking dataset, our FTT-spectral improves FTF-spectral upon around 3.5\% in ACC, 1.7\% in NMI, and 3.3\% in F1. And, on Finance dataset, 1.5\% in ACC, 0.3\% in NMI, and 1\% in F1. 

Comparing FTF and TTF, the general representations induced by GE module contribute to the performance. On Banking dataset, our TTF-spectral improves FTF-spectral upon around 6\% in ACC, 3\% in NMI, 6\% in F1. And, on Finance dataset, 2\% in ACC, 1\% in NMI, 1.5\% in F1. GE module is prominently helpful for banking dataset. We conjecture that Banking utterances are closer to ordinary language than Finance utterances.
Finally, TTF-spectral and TTT-spectral show that the combination of GE and PGT boosts ACC on Banking and Finance, around 0.5\% for the two datasets. 

In sum, simply changing from K-means to Spectral clustering improves the performance significantly in the model we submitted. Our TTT-spectral shows remarkable performance in two different domains (e.g., Banking, Finance) without additional training process. Mainly, the MDB method boosts performance by a large margin, and the PGT module raises the performance to a higher level. We empirically recognize that Spectral clustering shows better synergy with our models than K-means clustering. Furthermore, MDB and PGT methods work well regardless of the domain without additional fine-tuning processes.

\section{Conclusion}

\; In TOD systems, it is noteworthy to capture users' intents regardless of domain. We figure out that simply adopting PLM does not perform well in dialogue corpus. Therefore, PLM needs more training. We suggest the multi-view model with GE, MDB, and PGT modules. The MDB method with cosine softmax utilizes an existing dataset to bridge between PLM and dialogue domain. The novel PGT method, a methodology for fine-tuning based on K-means, is presented to enhance the clustering capability. As a result, the performance increases in Finance and Banking datasets. According to our clustering experiments, we found that Spectral clustering is best fitted for our model, resulting in more significant and higher performance than simple K-means methods.

\section*{Limitations}
\; The first limitation of our approach is that the hidden representation dimension increases as more views are used. Therefore, the clustering phase may take longer than other models. 
Second, interactive information among these multi-views is not considered in our model. Thus, we will introduce meta-attention to harmonize them, not just a concatenation, to reflect interactive information among modules. Lastly, in order to apply MDB, an existing task-related dataset must exist. Otherwise, it is impossible to train the MDB module.

\section*{Acknowledgement}
K. Jung is with ASRI, Seoul National University, Korea. This work was supported by Institute of Information \& communications Technology Planning \& Evaluation(IITP) grant funded by the Korea government(MSIT) [No. 2022-0-00184, Development and Study of AI Technologies to Inexpensively Conform to Evolving Policy on Ethics]. This work was partly supported by Institute of Information \& communications Technology Planning \& Evaluation (IITP) grant funded by the Korea government(MSIT) [NO.2021-0- 02068, Artificial Intelligence Innovation Hub (Artificial Intelligence Institute, Seoul National University) \& NO.2021-0-01343, Artificial Intelligence Graduate School Program (Seoul National University)]

\bibliography{anthology,custom}
\bibliographystyle{acl_natbib}

\appendix

\section{Kmeans vs Spectral in Banking Dataset}
\label{Banking appendix}

\subsection{Analysis about Banking Dataset}
\; Table \ref{simple-stats} shows some simple statistics for Banking Dataset. Some utterances do not belong to the gold intent sets, regarded as noises. As shown in the Tabel \ref{example-banking}, the utterances are colloquial and usually include filler words. In some cases, samples include utterances which are different from its label. Furthermore, some samples require context to fully capture their intention, such as "I need to make a deposit like now." Additionally, some utterances could be annotated with multiple labels. For example, "Can I close my savings account and transfer the remaining balance from my savings account into my checking account?", can be labeled as not only CloseBankAccount but also, InternalFundTransfer.

\; We used the utterances, including noise, to make an intent schema where 2,325 instances (63\%) were noise. They are daily conversations regardless of banking. Therefore, it is important to distinguish noise from other gold labels effectively. 

\begin{table*}
\centering
\caption{Simple statistics about Banking Dataset}
\label{simple-stats}
\begin{tabular}{|l|l|l|l}
\hline
Dataset file name             & \# of Utterances & \# of Intent & \multicolumn{1}{l|}{\# of Noise} \\ \hline
Dialogue.jsonl (InformIntent) & 3696             & 30           & \multicolumn{1}{l|}{2325}        \\ \hline
Test-utterance.jsonl          & 407              & 18           & \multicolumn{1}{l|}{0}           \\ \hline
\end{tabular}
\end{table*}

\begin{table*}
\centering
\caption{Examples of Samples in Banking Dataset}
\label{example-banking}
\begin{tabular}{|l|l|}
\hline
Utterances& Intent          \\ \hline
I need to make a deposit like now.& FindBranch      \\ \hline
No, can you help me find one please?& FindBranch      \\ \hline
\begin{tabular}[c]{@{}l@{}}Well I’ve got the mobile app. And I’ll probably just go ahead\\ and use that because I think I could get directions straight from it.\\ Yeah, I’m new to this new place, anyways. So what about fees?\\ What do they do with those cause I know you\\ when I lived in Alabama we had locations you\\ everywhere and I didn’t have to worry about those new fees\\ because I was always using you Intellibank’s ATMs\\ but so what what’s like the fees like if I if I don’t use an Intellibank?\end{tabular} & AskAboutATMFees \\ \hline
\begin{tabular}[c]{@{}l@{}}Can I close my savings account and transfer the remaining\\ balance from my savings account into my checking account?\end{tabular}& CloseBankAccoun \\ \hline
All right. So where are you from?& N/A             \\ \hline
All righty. So are are you in Raleigh?& N/A             \\ \hline
gonna take very long?  & N/A             \\ \hline
I just need the balance for my checking account, please. & N/A             \\ \hline
All right. That sounds great. Is is there a fee associated with this? & N/A             \\ \hline
And I need some money transferred to my daughter to pay her bills. & N/A             \\ \hline
\end{tabular}
\end{table*}

\; There is a huge data imbalance with respect to the number of instances included in each gold cluster; for example, the label "CheckAccountBalance" has 243 samples, while "AskAboutCreditScore" has only three samples.
The number of samples for each intent label is shown in Table \ref{goldcluster}. 

\begin{table*}[]
\centering
\caption{Number of samples included in each gold intent cluster in Banking}
\label{goldcluster}
\begin{tabular}{|l|l|l|l|}
\hline
AskAboutATMFees          & 8   & GetBranchHours        & 64   \\ \hline
AskAboutCardArrival      & 10  & GetBranchInfo         & 20   \\ \hline
AskAboutCashDeposits     & 15  & GetWithdrawalLimit    & 20   \\ \hline
AskAboutCreditScore      & 3   & InternalFundsTransfer & 130  \\ \hline
AskAboutTransferFees     & 11  & OpenBankingAccount    & 83   \\ \hline
AskAboutTransferTime     & 24  & OpenCreditCard        & 5    \\ \hline
CheckAccountBalance      & 243 & OrderChecks           & 9    \\ \hline
CheckAccountInterestRate & 10  & ReportLostStolenCard  & 70   \\ \hline
CheckTransactionHistory  & 23  & ReportNotice          & 17   \\ \hline
CloseBankAccount         & 63  & RequestNewCard        & 15   \\ \hline
DisputeCharge            & 89  & SetUpOnlineBanking    & 18   \\ \hline
ExternalWireTransfer     & 109 & UpdateEmail           & 34   \\ \hline
FindATM                  & 86  & UpdatePhoneNumber     & 20   \\ \hline
FindBranch               & 110 & UpdateStreetAddress   & 49   \\ \hline
GetAccountInfo           & 13  & SUM                   & 1371 \\ \hline
\end{tabular}
\end{table*}

\subsection{Analysis about Intent Schema and Test Prediction}
\; For noise detection, in the case of Kmeans clustering, this algorithm finds 28 intent clusters out of 30 gold clusters, whereas Spectral clustering found 34 intent clusters. Kmeans clustering detect 251 noise utterances out of 2325 noise instances and showed an F1 score of 0.189, while Spectral clustering found 838 noise instances and showed a significantly higher F1 score of 0.478. 
According to detected labels, the intent schema created through Kmeans clustering contained a lot of noise, resulting in an average F1 score of 0.288 without noise. In contrast, the intent schema created through Spectral clustering achieved a significantly higher average F1 score of 0.379 without noise. 

Spectral clustering presented a better performance for labeled data with noise. However, Kmeans clustering tended to concentrate on certain words and did not consider the meaning of the entire sentence. This makes it difficult to distinguish noise from the other gold labels and decide the boundaries between clusters, which share similar semantic with some variations such as "UpdateEmail" and "UpdateStreetAddress." In contrast, Spectral clustering tended to consider the meaning of the entire sentence, filtering out noises consisting of daily conversations. In fact, noises are labeled sentences within additional four clusters. 

\;Spectral clustering also showed better performance for labeled data without noise. 
Throughout table \ref{spectralvskmeans}, Spectral clustering clustered sentences which have similar meanings. It also appropriately handled unrelated utterances as NaN, while Kmeans clustering failed to do so. Analyzing the results among the labels, i.e., "UpdateEmail" and "UpdateStreetAddres," Spectral clustering considered the meaning of the entire sentence more, further accurately distinguishing noise from labeled utterances. We recognize that it detects seemingly insignificant word changes such as "email" and "street," preserving overall similar semantics. However, Kmeans clustering focused on specific words rather than the entire meaning of the sentence, resulting in misclassifying instances to noise or wrong label even if it contains the exact same keyword with the label. Moreover, their aligned utterances share semantic information; update something. Thus, they tend to locate nearby, obscuring boundaries in terms of the distance-based method. That is, Kmeans clustering cannot distinguish the subtle changes in meaning.

\; We applied the test-utterance dataset to the classifier trained on the Intent schema to conduct the final evaluation. As mentioned in the Results \ref{results}, the classifier trained on the schema created using spectral clustering showed remarkably high performance. When Spectral clustering algorithm creates the schema, it groups similar utterances discerning noises. On the other hand, Kmeans focused on specific words or misclassified daily conversations irrelevant to the topic, which lowered the quality of the schema and caused the classifier to make many errors. As a result, spectral clustering showed significantly higher performance in ACC. 

% Please add the following required packages to your document preamble:
% \usepackage[table,xcdraw]{xcolor}
% If you use beamer only pass "xcolor=table" option, i.e. \documentclass[xcolor=table]{beamer}
\begin{table*}[]
\centering
\scriptsize
\caption{Examples about classifiers' predicted labels in Banking}
\label{spectralvskmeans}
\begin{tabular}{|l|l|l|l|}
\hline
\toprule Utterance & Gold intent & Predicted label (Kmeans) & Predicted label (Spectral)\\ \hline
And and is it. Is it free? Is it free?                 & NaN & AskAboutATMFees       & AskAboutATMFees     \\ \hline
Oh, thanks. What about you? Are you having weather?    & NaN & AskAboutCardArrival   & NaN                 \\ \hline
Do you guys have an app?                               & NaN & SetUpOnlineBanking    & NaN                 \\ \hline
\begin{tabular}[c]{@{}l@{}}OK. that's what what what about 
my my savings account.\end{tabular} & NaN & InternalFundsTransfer & CheckAccountBalance \\ \hline
I want to put money into his account.	&InternalFundsTransfer	&ExternalWireTransfer	&InternalFundsTransfer \\ \hline
\begin{tabular}[c]{@{}l@{}}no, if possible, could I update my email with you?\\ That's an old one.\end{tabular}	&UpdateEmail	&UpdateStreetAddress	&UpdateEmail \\ \hline
\begin{tabular}[c]{@{}l@{}}OK. I appreciate that.\\ And I wanted to also update my number.\end{tabular}	&UpdatePhoneNumber	&UpdateStreetAddress	&UpdatePhoneNumber \\ \hline
\begin{tabular}[c]{@{}l@{}}... I have money taken of my account \\ that I didn't that I did not do.\end{tabular}	&DisputeCharge	&CheckTransactionHistory	&DisputeCharge \\ \hline
\end{tabular}
\end{table*}

\subsection{Alignment}

\;In the organizer's code, predicted labels are assigned one-to-one with reference labels. The one-to-one assignment is the Hungarian mapping method relying on the number of references, so duplicate allocation to the same reference is not allowed.
Therefore, when predicted clusters (23 labels) exceed the number of gold clusters (18 labels), almost similar clusters are divided into two different intents, one of which is abandoned. If we aligned the reference label to the predicted cluster allowing overlapping, both Kmeans and Spectral results show that ACC increased by about 4.5\% (0.786 and 0.953, respectively). Spectral F1 score also increased by 2\%, while Kmeans F1 score raised by 0.1\%. It indicates that spectral clustering was well-trained in sensitivity to noise and effectively separated existing clusters.

\section{Kmeans vs Spectral in Finance Dataset}
\label{Finance appendix}
\begin{table*}
    \centering
    \caption{Kmeans clustering's incorrect predictions and representative examples in Finance. }
    \label{finance-kmeans}
    \scriptsize
    
    \begin{tabular}{|l|l|l|}
    \toprule
    Predict  & Reference & utter \\
    \midrule
    AddUserToAccount           & ScheduleAppointment        & Good morning. I need to \textbf{see a notary} for some documents by at least 2pm today please.        \\
    CancelCheck      & OrderCheck       & I need to order \textbf{checks} ASAP! \\
    CancelCheck      & OrderCheck       & I want to order 500 \textbf{checks} for my business. \\
    ChangePin       & CloseAccount       & please close account \#098716253 with \textbf{pin} \#65491       \\
    ChangePin          & GetCreditReport       & ... I need my business credit report sent to me via mail. My \textbf{pin} is 94442        \\
    UpdateEmail           & UpdatePhoneNumber        & I'd like to \textbf{update} my phone number is 761-451-9850        \\
    UpdateEmail   & UpdateStreetAddress        & I'd like to \textbf{update} my street address please, my name is Helen Smithfield        \\
    PurchaseStocks   & AskLiquidityRatio        & i want to check \textbf{liquidity} ratio        \\
    PurchaseStocks   & AskLiquidityRatio        & i am jorah mont, i want to check my \textbf{liquidity} ratio        \\
    OpenAccount   & CloseAccount        & Hi please close my \textbf{account}, I'm switching to a different \textbf{bank}        \\
    \bottomrule
\end{tabular}
\end{table*}
\begin{table*}
    \centering
    \caption{Spectral clustering's top incorrect prediction-label pairs in Finance.}
    \label{finance-spectral-pair}
    \begin{tabular}{|l|l|}
    \toprule
    Predict  & Reference \\
    \midrule
    ApplyCreditCard           & RequestNewCard      \\
    checkAccountBalance      & CheckCreditCardBalance       \\
    MakeCreditCardPayment      & SetAutoPayment      \\
    FindBranch       & ScheduleAppointment      \\
    \bottomrule
\end{tabular}
\end{table*}
\begin{table*}
    \centering
    \caption{Spectral clustering's incorrect predictions and representative examples in Finance.}
    \label{finance-spectral}
    \scriptsize
    \begin{tabular}{|l|l|l|}
    \toprule
    Predict  & Reference & utter \\
    \midrule
    GetBranchHours           & FindBranch        & What's the nearest branch I can visit in person?        \\
    FindBranch      & ScheduleAppointment       & I need to see a person about my taxes. In person thank you. \\
    GetBranchHours      & FindBranch       & Which of your branches can I visit in person? \\
    UpdateEmail       & ChangeStatementDelivery       & hey I want to \textbf{change} my monthly statement delivery to \textbf{email} pls       \\
    UpdatePhoneNumber          & AddUserToAccount       & ... I need to add Luke Walker to my business account. The \textbf{account number} is 45000222        \\
    UpdateStreetAddress & AddUserToAccount & I'm Eric Daniels. I need to add \textbf{Gerald Smith} to my business account. \\
    \bottomrule
\end{tabular}
\end{table*}

\; This section presents an empirical analysis of the results obtained from "TTT-kmeans" and "TTT-spectral" on the finance dataset. The Kmeans clustering tends to prioritize specific keywords over the overall context. For instance, in the first example provided in Table \ref{finance-kmeans}, taking a look at the bold phrases, it can infer "ScheduleAppointment" from "see a notary," but may provide incorrect answers if it fails to comprehend such chunks. Moreover, in cases such as "OrderCheck," where it encounters the word "check," it tends to blindly map it to "CancelCheck," and for sentences containing the word "pin," it predicts "ChangePin." Additionally, when it incorrectly predicts "UpdateEmail," it directly maps the word "update" to "UpdateEmail," and there is a tendency to link the word "liquidity" to "purchase."

 On the other hand, spectral clustering tends to focus less on specific keywords than Kmeans clustering, but may still make errors in distinguishing relatively ambiguous labels in Table \ref{finance-spectral-pair}. In cases where it makes incorrect predictions, there are often instances of multi-labeling depending on how to interpret utterances. For example, in table \ref{finance-spectral}, the sentence "what's the nearest branch I can visit in person?" carries an ambiguous meaning, such as where I can go and which branch is currently open. However, given the erroneous mapping of "change statement" to "UpdateEmail" and the misinterpretation of "add to business account" as "PhoneNumber" due to its focus on the word "number," and the misclassification of the entity "Gerald Smith" as a location name, there is still much room for improvement.

\end{document}